\documentclass[10pt,twocolumn,letterpaper]{article}

\usepackage{cvpr}
\usepackage{times}
\usepackage{epsfig}
\usepackage{graphicx}
\usepackage{amsmath}
\usepackage{amssymb}
\usepackage{booktabs}
\usepackage{multirow}
\usepackage{epstopdf}
\usepackage{nomencl}

\usepackage{rotating}
\usepackage{subcaption}
\usepackage{array}
\usepackage{url}

\usepackage[pagebackref=true,breaklinks=true,letterpaper=true,colorlinks,bookmarks=false]{hyperref}

\cvprfinalcopy % *** Uncomment this line for the final submission

\def\cvprPaperID{741} % *** Enter the CVPR Paper ID here

\ifcvprfinal\fi
\begin{document}

%%%%%%%%% TITLE
\title{Learning a Deep Embedding Model for Zero-Shot Learning}

\author{Li Zhang \quad  Tao Xiang \quad  Shaogang Gong\\
Queen Mary University of London\\
{\tt\small \{david.lizhang, t.xiang, s.gong\}@qmul.ac.uk}
}

\maketitle
\thispagestyle{empty}

%%%%%%%%% ABSTRACT
\begin{abstract}
Zero-shot learning (ZSL) models rely on learning a joint embedding space where both textual/semantic description of object classes and visual representation of object images can be projected to for nearest neighbour search. Despite the success of deep neural networks that learn an end-to-end model between text and images in other vision problems such as image captioning, very few deep ZSL model exists and they show little advantage over ZSL models that utilise deep feature representations but do not learn an end-to-end embedding. In this paper we argue that the key to make deep ZSL models succeed is to choose the right embedding space. Instead of embedding into a semantic space or an intermediate space, we propose to use the visual space as the embedding space. This is because that in this space, the subsequent nearest neighbour search would suffer much less from the hubness problem and thus become more effective. This  model design also provides a natural mechanism for  multiple semantic modalities (e.g.,~attributes and sentence descriptions) to be fused and optimised jointly in an end-to-end manner. Extensive experiments on four benchmarks show that our model significantly outperforms the existing models. Code is available at: \url{https://github.com/lzrobots/DeepEmbeddingModel_ZSL}

\end{abstract}

%%%%%%%%% BODY TEXT
\section{Introduction}

A recent trend in developing visual recognition models is to scale up the number of object categories.  However, most existing recognition models are based on supervised learning and  require a large amount (at least 100s) of training samples to be collected and annotated for each object class to capture its intra-class appearance variations \cite{deng2009imagenet}. This severely limits their scalability -- collecting daily objects such as chair is easier, but many other categories are rare ({\em e.g.}, a  newly identified  specie of beetle on a remote pacific island). None of these models can work with few or even no training samples for a given class. In contrast, humans are very good at recognising objects without seeing any visual samples, i.e., {\em zero-shot learning} (ZSL). For example, a child would have no problem recognising a zebra if she has seen horses before and {\em also} read elsewhere that a zebra is a horse but with black-and-white stripes on it. Inspired by humans' ZSL ability, recently there is a surge of interest in machine ZSL~\cite{akata2015evaluation, zhang2015zero, lampert2014attribute, akata2013label, romera2015embarrassingly, socher2013zero, frome2013devise, norouzi2013zero, fu2014transductive, fu2015zero, lei2015predicting, yang2014unified, reed2016learning, changpinyo2016synthesized, fu2016semi, bucher2016improving, chao2016empirical, zhang2016zero, zhang2016zeroshot}. 

A zero-shot learning method relies on the existence of a labelled training set of {\em seen classes} and the knowledge about how an {\em unseen class} is semantically related to the seen classes. Seen and unseen classes are usually related in a high dimensional vector space, called semantic space, where the knowledge from seen classes can be transferred to unseen classes. The semantic  spaces used by most early works are based on semantic attributes \cite{farhadi2009describing, ferrari2007learning, parikh2011relative}. Given a defined attribute ontology, each class name can be represented by an attribute vector and termed as a class {\em prototype}. More recently,  semantic word vector space \cite{socher2013zero, frome2013devise} and  sentence descriptions/captions \cite{reed2016learning} have started to gain popularity. With the former, the class names are projected into a word vector space so that different classes can be compared, whilst with the latter, a neural language model is required to provide a vector representation of the description. 
 
With the semantic space and a visual feature representation of image content, ZSL is typically solved in two steps: (1) A joint embedding space is learned where both the semantic vectors (prototypes) and the visual feature vectors can be projected to; and (2) nearest neighbour (NN) search is performed in this embedding space to match the projection of an image feature vector against that of an unseen class prototype. Most state-of-the-arts ZSL models \cite{fu2014transductive, fu2016semi, akata2015evaluation, bucher2016improving, romera2015embarrassingly, zhang2015zero, lampert2014attribute} use deep CNN features for visual feature representation; the features are extracted with pretrained  CNN models. They differ mainly in how to learn the embedding space given the features. They are thus not end-to-end deep learning models.

%why deep: (1) end-to-end learning, in particular the text modality. (2) flexibity for multi-modality and multiple tranfer learning per Yongxing's paper, (3) flexible for various tasks such as retrieval and even genreate novel description of novel class. 
In this paper, we focus on end-to-end learning of a deep embedding based ZSL model which offers a number of advantages. First, end-to-end optimisation can potentially lead to learning a better embedding space. For example, if sentence descriptions are used as the input to a neural language model such as recurrent neural networks (RNNs) for computing a semantic space, both the neural language model and the CNN visual feature representation learning model can be jointly optimised in an end-to-end fashion. Second, a neural network based joint embedding model offers the flexibility for addressing various transfer learning problems such as multi-task learning and multi-domain learning \cite{yang2014unified}. Third, when multiple semantic spaces are available, this model can provide a natural mechanism for fusing the multiple modalities. However, despite  all these intrinsic advantages, in practice, the few existing end-to-end deep models for ZSL in the literature \cite{lei2015predicting, frome2013devise, socher2013zero, yang2014unified, reed2016learning} fail to demonstrate these
advantages and yield only weaker or merely comparable performances on benchmarks when compared to non-deep learning alternatives.

We argue that the key to the success of a deep embedding model for ZSL is the choice of the embedding space. Existing models, regardless whether they are deep or non-deep, choose either the semantic space  \cite{lampert2014attribute, fu2016semi,socher2013zero, frome2013devise} or an intermediate embedding space  \cite{lei2015predicting, akata2015evaluation, romera2015embarrassingly, fu2014transductive}  as the embedding space. However, since the embedding space is of high dimension and NN search is to be performed there, the hubness problem is inevitable  \cite{radovanovic2010hubs}, that is, a few unseen class prototypes will become the NNs of many data points, i.e., hubs. Using the semantic space as the embedding space means that the visual feature vectors need to be projected into the semantic space which will shrink the variance of the projected data points and thus aggravate the hubness problem \cite{radovanovic2010hubs,dinu2014improving}. 

In this work, we propose a novel Deep neural network based Mmbedding Model (DEM) for ZSL which differs from existing models in that: (1) To alleviate the hubness problem, we use the output visual feature space of a CNN subnet as the embedding space. The resulting projection direction is from a semantic space, e.g., attribute or word vector, to a visual feature space. Such a direction is opposite to the one adopted by most existing models. We provide a theoretical analysis and some intuitive visualisations to explain why this would help us counter the hubness problem. (2) A simple yet effective multi-modality fusion
method is developed in our neural network model which is flexible and importantly enables end-to-end learning of the semantic space representation.

The contributions of this work are as follows: (i) A novel deep embedding model for ZSL has been formulated which differs from existing models in the selection of embedding space. (ii) A multi-modality fusion method is further developed to combine different semantic representations and to enable end-to-end learning of the representations.  
Extensive experiments carried out on four benchmarks including AwA \cite{lampert2014attribute}, CUB \cite{wah2011multiclass} and large scale ILSVRC 2010 and ILSVRC 2012~\cite{deng2009imagenet} show that our model beats all the state-of-the-art models presented to date, often by a large margin.

%-------------------------------------------------------------------------
\section{Related Work}
\label{sec:related work}
\noindent \textbf{Semantic space} \quad Existing ZSL methods differ in what semantic spaces are used: typically either  attribute~\cite{farhadi2009describing, ferrari2007learning, parikh2011relative, sung2018learning}, word vector ~\cite{socher2013zero, frome2013devise}, or text description~\cite{reed2016learning, zhang2017actor}. It has been shown  that an attribute space is often more effective than a word vector space \cite{akata2015evaluation, zhang2015zero, lampert2014attribute, romera2015embarrassingly}. This is hardly surprising as additional attribute annotations are required for each class. Similarly, state-of-the-art results on fine-grained recognition tasks have been achieved in ~\cite{reed2016learning} using image sentence descriptions to construct the semantic space. Again, the good performance is obtained at the price of more manual annotation: 10 sentence descriptions need to be collected for each image, which is even more expensive than attribute annotation. This is why the word vector semantic space is still attractive: it is `free' and is the only choice for large scale recognition with many unseen classes \cite{fu2016semi}. In this work, all three semantic spaces are considered.

\noindent \textbf{Fusing multiple semantic spaces } \quad Multiple semantic spaces are often complementary to each other; fusing them thus can potentially lead to improvements in recognition performance. Score-level fusion is perhaps the simplest strategy \cite{fu2015zero}. More sophisticated multi-view embedding models have been proposed. Akata et al.~\cite{akata2015evaluation} learn a joint embedding semantic space between attribute, text and hierarchical relationship which relies heavily on hyperparameter search. Multi-view canonical correlation  analysis (CCA) has also been employed \cite{fu2014transductive} to explore different modalities of testing data in a transductive way. Differing from these models, our neural network based model has an embedding layer to fuse different semantic spaces and connect the fused representation with the rest of the visual-semantic embedding network for end-to-end learning. Unlike  \cite{fu2014transductive}, it is inductive and does not require to access the whole test set at once.

\noindent \textbf{Embedding model} \quad Existing methods also differ in the visual-semantic embedding model used. They  can be categorised into two groups: (1) The first group learns a mapping function  by regression from the visual feature space to the semantic space with pre-computed features ~\cite{lampert2014attribute, fu2016semi} or deep neural network regression~\cite{socher2013zero, frome2013devise}. For these embedding models, the semantic space is the embedding space. (2) The second group of models implicitly learn the relationship between the visual and semantic space through  a common intermediate space,  again either with a neural network formulation \cite{lei2015predicting, yang2014unified} or without \cite{lei2015predicting, akata2015evaluation, romera2015embarrassingly, fu2014transductive}. The embedding space is thus neither the visual feature space, nor the semantic space.  We show in this work that using the visual feature space as the embedding space is intrinsically advantageous due to its ability to alleviate the hubness problem.

\noindent \textbf{Deep ZSL model} \quad
All recent ZSL models use deep CNN features as inputs to their embedding model. However, few are deep end-to-end models. Existing deep neural network based ZSL works~\cite{frome2013devise, socher2013zero, lei2015predicting, yang2014unified,reed2016learning} differ in whether they use the semantic space or an intermediate space as the embedding space, as mentioned above.  They also use different losses. Some of them use margin-based losses \cite{frome2013devise, yang2014unified, reed2016learning}. Socher {\em et al} \cite{socher2013zero} choose a euclidean distance loss. Ba {\em et al} \cite{lei2015predicting} takes a dot product between the embedded visual feature and semantic vectors and consider three training losses, including a binary cross entropy loss, hinge loss and Euclidean distance loss.  In our model, we find that the least square loss between the two embedded vectors is very effective and offers an easy theoretical justification as for why it copes with the hubness problem better. 
The work in \cite{reed2016learning} differs from the other models in that it integrates a neural language model into its neural network for end-to-end learning of the embedding space as well as the language model. In additional to the ability of jointly learning the neural language model and embedding model, our model is capable of fusing text description with other semantic spaces and achieves better performance than \cite{reed2016learning}.

\noindent{\bf The hubness problem} \quad The phenomenon of the presence of
`universal' neighbours, or hubs, in a high-dimensional space for
nearest neighbour search was first studied by Radovanovic et al.~\cite{marcobaronihubness}. They show that hubness is an inherent
property of data distributions in a high-dimensional vector space, and
a specific aspect of the curse of dimensionality. A couple of recent
studies \cite{dinu2014improving,shigeto2015ridge} noted that regression based 
zero-shot learning methods suffer from the hubness problem and
proposed solutions to mitigate the hubness problem. Among them, the
method in \cite{dinu2014improving} relies on the modelling of the
global distribution of test unseen data ranks w.r.t. each class
prototypes to ease the hubness problem. It is thus transductive. In
contrast, the method in \cite{shigeto2015ridge} is inductive: It argued
that least square regularised projection functions make the hubness
problem worse and proposed to perform reverse regression, i.e.,
embedding class prototypes into the visual feature space. Our model also uses the visual feature space as the embedding space but achieve so by using an end-to-end deep neural network which yields far superior performance on ZSL.

\section{Methodology}
\subsection{Problem definition}
\label{Problem}  

Assume a labelled training set of $N$ training samples is given as $\mathcal{D}_{tr} = \{(\mathbf{I}_{i}, ~\mathbf{y}^{u}_{i}, ~t_{i}^{u}), i = 1, \dots , N \}$, with associated class label set $\mathcal{T}_{tr}$, where $\mathbf{I}_{i}$ is the $i$-th training image, $\mathbf{y}^{u}_{i} \in \mathbb{R}^{L \times 1}$ is its corresponding $L$-dimensional semantic representation vector, $t_{i}^{u} \in \mathcal{T}_{tr}$ is the $u$-th training class label for the $i$-th training image.  Given a new test image $\mathbf{I}_{j}$, the goal of ZSL is to predict a class label $t_{j}^{v} \in \mathcal{T}_{te}$, where $t_{j}^{v}$ is the $v$-th test class label for the $j$-th test instance. We have 
$\mathcal{T}_{tr} \cap \mathcal{T}_{te} = \varnothing$, i.e., the training (seen) classes and test (unseen) classes are disjoint. Note that each class label $t^{u}$ or $t^{v}$ is associated with a pre-defined semantic space representation $\mathbf{y}^{u}$ or $ \mathbf{y}^{v}$ (e.g.~attribute vector), referred to as semantic class prototypes. For the training set, $\mathbf{y}^{u}_{i}$ is given because each training  image $\mathbf{I}_{i}$ is labelled by a semantic representation vector representing its corresponding class label $t_j^{u}$. %Two ZSL settings are considered: 

\subsection{Model architecture}

		\begin{figure*}
	\centering
		\includegraphics[height=5.67cm]{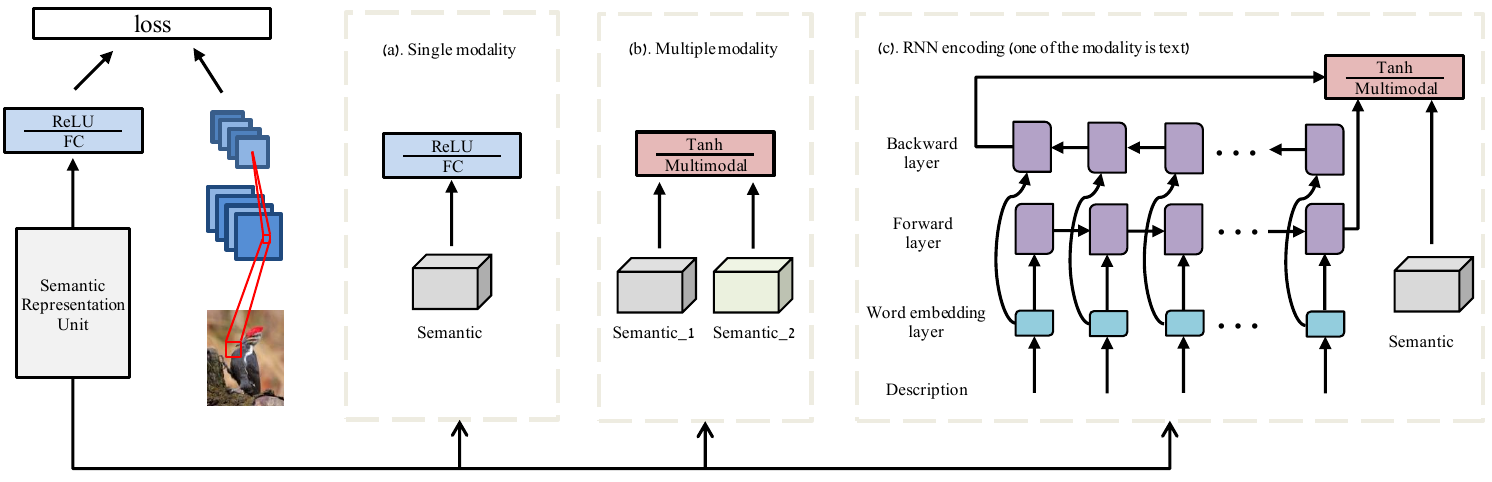}
		%\vspace{-0.3cm}
		\caption{Illustration of the network architecture of our deep embedding model. The detailed architecture of the semantic representation unit in the left branch (semantic encoding subnet) is given in (a), (b) and (c) which correspond to the single modality (semantic space) case, the multiple (two) modality case, and the case where one of the modalities is text description. For the case in (c), the semantic representation itself is a neural network (RNN) which is learned end-to-end with the rest of the network. 
		}
		\label{fig:full_model}
	\end{figure*} 

The architecture of our model is shown in Fig.~\ref{fig:full_model}. It has two branches. One branch is the visual encoding branch, which consists of a CNN subnet that takes an image $\mathbf{I}_{i}$ as input and outputs a $D$-dimensional feature vector $\phi(\mathbf{I}_{i})\in \mathbb{R}^{D \times 1}$. This $D$-dimensional visual feature space will be used as the embedding space where both the image content and the semantic representation of the class that the image belongs to will be embedded. The semantic embedding is achieved by the other branch which is a semantic encoding subnet. Specifically, it takes a $L$-dimensional semantic representation vector of the corresponding class $\mathbf{y}^{u}_{i}$ as input, and after going through two fully connected (FC) linear  + Rectified Linear Unit (ReLU) layers outputs a $D$-dimensional semantic embedding vector. Each of the FC layer has a $l_{2}$ parameter regularisation loss. The two branches are linked together by a least square embedding loss which aims to minimise the discrepancy between the visual feature $\phi(\mathbf{I}_{i})$ and its class representation embedding vector in the visual feature space. With the three losses, our objective function is as follows:

\begin{eqnarray}\label{loss_function}   
\mathcal{L}(\mathbf{W}_1, \mathbf{W}_2) = \frac{1}{N}\sum^{N}_{i=1} ||\phi(\mathbf{I}_{i}) - f_{1}(\mathbf{W}_{2}f_{1} (\mathbf{W}_{1}\mathbf{y}^{u}_{i})) ||^2 \nonumber \\ + \lambda( ||\mathbf{W}_1||^2+||\mathbf{W}_2||^2)
\end{eqnarray}
where $\mathbf{W}_1 \in \mathbb{R}^{L \times M}$ are the weights to be learned in the first FC layer and $\mathbf{W}_2 \in \mathbb{R}^{M \times D}$ for the second FC layer. $\lambda$ is the hyperparameter weighting the strengths of the two parameter regularisation losses against the embedding loss. We set $f_{1}(\centerdot)$ to be the Rectified Linear Unit (ReLU) which introduces nonlinearity in the encoding subnet~\cite{krizhevsky2012imagenet}. 

%\noindent
After that, the classification of the test image $\mathbf{I}_{j}$ in the visual feature space can be achieved by simply calculating its distance to the embed prototypes:

\begin{equation} 
v = \arg\min_{v} \mathcal{D} (\phi(\mathbf{I}_{j}), f_{1}(\mathbf{W}_{2}f_{1} (\mathbf{W}_{1}\mathbf{y}^{v})))
\end{equation} 
\noindent
where $\mathcal{D}$ is a distance function, and $\mathbf{y}^{v}$ is the semantic space vector of the $v$-th test class prototype.

\subsection{Multiple semantic space fusion}
\label{sec:fusion}
As shown in Fig.~\ref{fig:full_model}, we can consider the semantic representation and the first FC and ReLU layer together as a semantic representation unit. When there is only one semantic space considered, it is illustrated in Fig.~\ref{fig:full_model}(a). However, when more than one semantic spaces are used, e.g., we want to fuse attribute vector with word vector for semantic representation of classes, the structure of the semantic representation unit is changed slightly, as shown in Fig.~\ref{fig:full_model}(b).

More specifically, we map different semantic representation vectors to a  multi-modal fusion layer/space where they are added. The output of the semantic representation unit thus becomes:

\begin{equation}\label{multimodal}    
f_{2}(\mathbf{W}_1^{(1)} \cdot \mathbf{y}^{u_1}_{i} + \mathbf{W}_1^{(2)} \cdot \mathbf{y}^{u_2}_{i} ),
\end{equation}
where $\mathbf{y}^{u_1}_{i} \in \mathbb{R}^{L_1 \times 1}$ and $\mathbf{y}^{u_2}_{i} \in \mathbb{R}^{L_2 \times 1}$ denote two different semantic space representations (e.g., attribute and word vector), ``+'' denotes element-wise sum,  $\mathbf{W}_1^{(1)}\in \mathbb{R}^{L_1 \times M}$ and $\mathbf{W}_1^{(2)}\in \mathbb{R}^{L_2 \times M}$ are the weights which will be learned. $f_{2}(\centerdot)$ is the element-wise scaled hyperbolic tangent function~\cite{lecun2012efficient}:
\begin{equation}\label{tanh}    
f_{2}(x) = 1.7159 \cdot \mathrm{tanh}(\frac{2}{3}x).
\end{equation}

This activation function forces the gradient into the most non-linear value range and leads to a faster training process than the basic hyperbolic tangent function.

\subsection{Bidirectional LSTM encoder for description}

The structure of the semantic representation unit needs to be changed again, when text description is avalialbe for each training image (see  Fig.~\ref{fig:full_model}(c)). In this work, we use a recurrent neural network (RNN) to encode the content of a text description (a variable length sentence) into a fixed-length semantic vector. Specifically,  given a text description of $T$ words,  $x = (x_{1}, \dots , x_{T})$  we use a Bidirectional RNN model \cite{schuster1997bidirectional} to encode them. For the RNN cell, the Long-Shot Term Memory (LSTM) \cite{hochreiter1997long} units are used as the recurrent units. The LSTM is a special kind of RNN, which introduces the concept of gating to control the message passing between different times steps. In this way, it could potentially model long term dependencies. Following  \cite{lstm_cell}, the model has two types of states to keep track of the historical records: a cell state $\textbf{c}$ and a hidden state $\textbf{h}$. For a particular time step $t$, they are computed by integrating the current inputs $x_t$ and previous state $(\textbf{c}_{t-1},\textbf{h}_{t-1})$. During the integrating, three types of gates are used to control the messaging passing: an input gate $\textbf{i}_t$, a forget gate $\textbf{f}_t$ and an output gate $\textbf{o}_t$.  

We omit the formulation of the bidirectional LSTM here and refer the readers to \cite{lstm_cell, graves2013hybrid} for details. With the bidirectional LSTM model, we use the final output as our encoded semantic feature vector to represent the text description:

\begin{equation}\label{bidirectionalLSTM}    
f(\mathbf{W}_{\overrightarrow{\mathbf{h}}} \cdot \overrightarrow{\mathbf{h}} + \mathbf{W}_{\overleftarrow{\mathbf{h}}} \cdot \overleftarrow{\mathbf{h}} ),
\end{equation}
where $\overrightarrow{\mathbf{h}}$ denote the forward final hidden state, $\overleftarrow{\mathbf{h}}$ denote the backward final hidden state. $f(\centerdot) = f_{1}(\centerdot)$ if text description is used only for semantic space unit, and
$f(\centerdot) = f_{2}(\centerdot)$ if other semantic space need to be fused (Sec. \ref{sec:fusion}). $\mathbf{W}_{\overrightarrow{\mathbf{h}}}$ and $\mathbf{W}_{\overleftarrow{\mathbf{h}}}$ are the weights which will be learned.

In the testing stage, we first extract text encoding from test descriptions and then average them per-class to form the test prototypes as in \cite{reed2016learning}. Note that since our ZSL model is a neural network, it is possible now to learn the RNN encoding subnet using the training data together with the rest of the network in an end-to-end fashion. 

\subsection{The hubness problem}
\label{sec:hubness}
How does our model deal with the hubness problem? First we show that our objective function is closely related to that of the ridge regression formulation. In particular, if we use the matrix form and write the outputs of the semantic representation unit as $\mathbf{A}$ and the outputs of the CNN visual feature encoder as $\mathbf{B}$, and ignore the ReLU unit for now,  our training objective becomes  
\begin{equation}\label{ridge_regression}   
\mathcal{L}(\mathbf{W}) = ||\mathbf{B} - \mathbf{W} \mathbf{A}||^2_{F} + \lambda ||\mathbf{W} ||_{F}^{2},
\end{equation}
which is basically ridge regression. It is well known that ridge regression has a closed-form solution $\mathbf{W} = \mathbf{B}\mathbf{A}^{\top}(\mathbf{A}\mathbf{A}^{\top} + \lambda \mathbf{I})^{-1}$. Thus we have:

\begin{eqnarray}\label{ridge_regression2}   
|| \mathbf{W}\mathbf{A}||_{2} = ||\mathbf{B}\mathbf{A}^{\top}(\mathbf{A}\mathbf{A}^{\top} + \lambda \mathbf{I})^{-1} \mathbf{A}||_{2} \nonumber \\ \leq ||\mathbf{B}||_{2}||\mathbf{A}^{\top}(\mathbf{A}\mathbf{A}^{\top} + \lambda \mathbf{I})^{-1} \mathbf{A}||_{2}
\end{eqnarray}

It can be further shown that 
\begin{equation}\label{ridge_regression3}   
||\mathbf{A}^{\top}(\mathbf{A}\mathbf{A}^{\top} + \lambda \mathbf{I})^{-1} \mathbf{A}||_{2} = \frac{\sigma^2}{\sigma^2 + \lambda} \leq 1.
\end{equation}
Where $\sigma$ is the largest singular value of $\mathbf{A}$. So we have $|| \mathbf{W}\mathbf{A}||_{2} \leq ||\mathbf{B} ||_{2}$.  This means the mapped source data $||\mathbf{WA}||_{2}$ are likely to be closer to the origin of the space than the target data $||\mathbf{B} ||_{2}$, with a smaller variance.

\begin{figure}[h!]
\centering
\begin{tabular}{cc}
{\includegraphics[width=4cm]{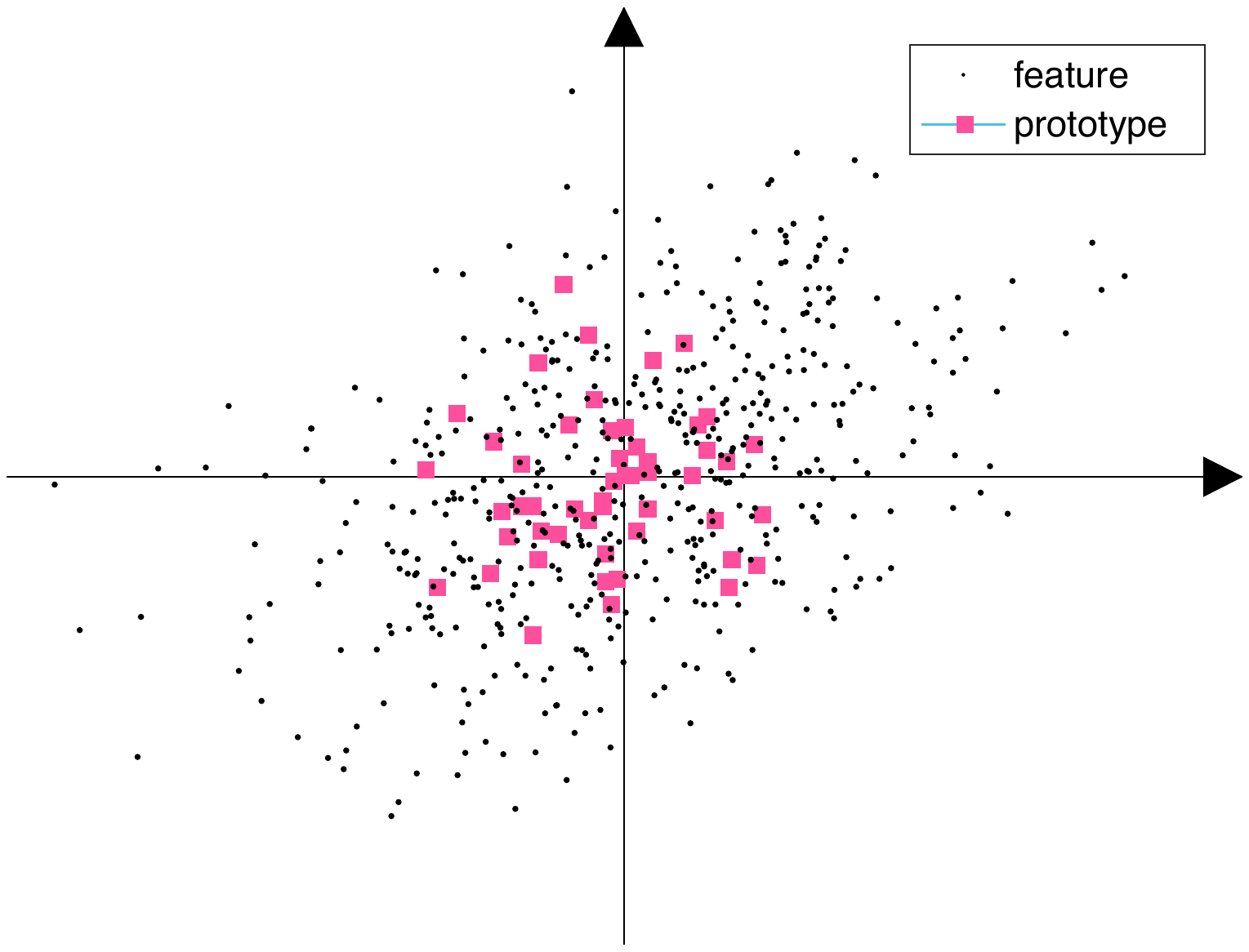}}&
{\includegraphics[width=4cm]{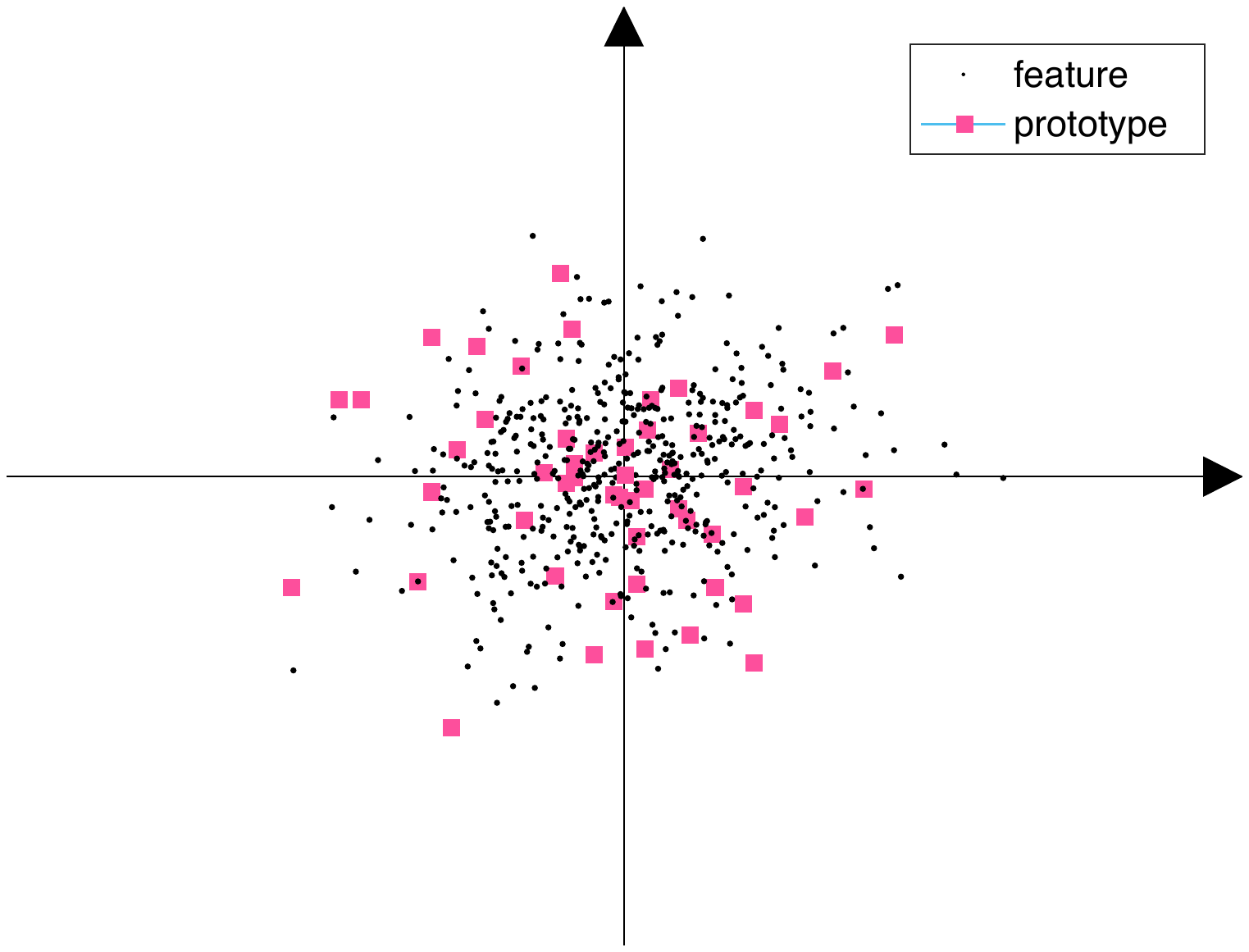}}\\
(a) S $\rightarrow$ V &(b) V $\rightarrow$ S
\end{tabular}
\vspace{0.3cm}
\caption{Illustration of the effects of different embedding directions on the hubness problem. S: semantic space, and V: visual feature space. Better viewed in colour.}
\label{fig:hubness}
\end{figure}

Why does this matter in the context of ZSL? Figure \ref{fig:hubness} gives an intuitive explanation. Specifically, assuming the feature distribution is uniform in the visual feature space, Fig.~\ref{fig:hubness}(a) shows that if the projected class prototypes are slightly shrunk towards the origin, it would not change how hubness problem arises -- in other words, it at least does not make the hubness issue worse. However, if the mapping direction were to be reversed, that is, we use the semantic vector space as the embedding space and project the visual feature vectors $\phi (\mathbf{I})$ into the space, the training  objective is still ridge regression-like, so the projected visual feature representation vectors will be shrunk  towards the origin as shown in Fig.~\ref{fig:hubness}(b). Then there is an adverse effect: the semantic vectors which are closer to the origin are more likely to  become hubs, i.e.~nearest neighbours to many projected visual feature representation vectors. This is confirmed by our experiments (see Sec.~\ref{sec:exp}) which show that using which space as the embedding space makes a big difference in terms of the degree/seriousness of the resultant hubness problem and therefore the ZSL performance.

\noindent {\bf Measure of hubness}\quad To measure the degree of hubness in a nearest neighbour search problem,  the {\em skewness} of the (empirical) $N_{k}$ distribution is used, following ~\cite{radovanovic2010hubs, shigeto2015ridge}. The $N_{k}$ distribution is the distribution of the number $N_{k}(i)$ of times each prototype $i$ is found in the top $k$ of the ranking for test samples (i.e.~their $k$-nearest neighbour), and its skewness is defined as follows:
\begin{equation} 
(N_{k} skewness) = \frac{\sum^{l}_{i=1}(N_{k}(i)-E[N_{k}])^{3}/l}{Var[N_{k}]^{\frac{3}{2}}},
\end{equation}
 where $l$ is the total number of test prototypes. A large  {\em skewness} value indicates the emergence of more hubs.

\subsection{Relationship to other deep ZSL models}
\begin{figure}[h]
\centering
\begin{tabular}{cc}
{\includegraphics[width=3.9cm]{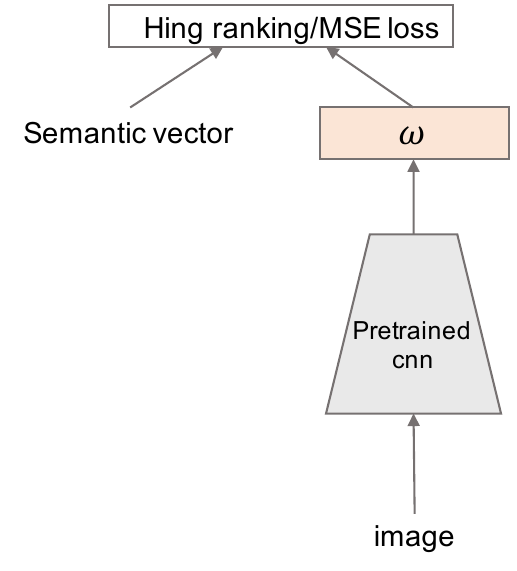}}&
{\includegraphics[width=3.9cm]{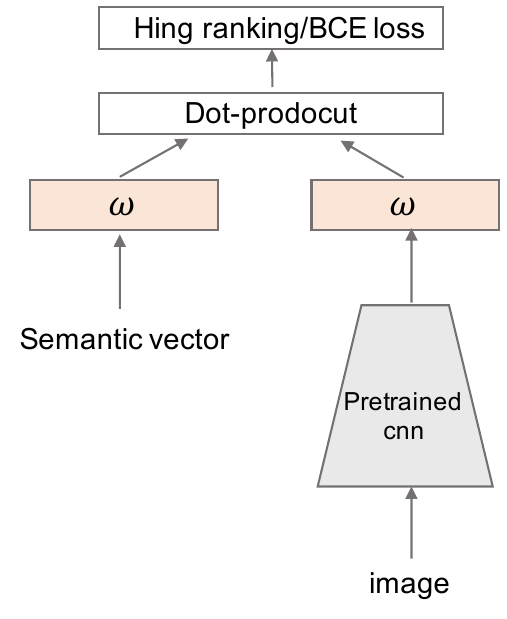}}\\
(a)~\cite{frome2013devise, socher2013zero} &(b)~\cite{yang2014unified, lei2015predicting}
\end{tabular}
\vspace{0.3cm}
\caption{The architectures of existing deep ZSL models fall into two groups: (a) learning projection function $\omega$ from visual feature space to semantic space; (b) learning an intermediate space as embedding space.}
\label{analysis1}
\end{figure}

Let's now compare the proposed model  with the related end-to-end neural network based models: DeViSE~\cite{frome2013devise}, Socher {\em et al.}~\cite{socher2013zero}, MTMDL~\cite{yang2014unified}, and Ba {\em et al.}~\cite{lei2015predicting}. Their model structures fall into two groups. In the first group (see Fig.~\ref{analysis1}(a)),  DeViSE~\cite{frome2013devise} and Socher {\em et al.}~\cite{socher2013zero} map the CNN visual feature vector to a semantic space by a hinge ranking loss or least square loss. In contrast,  MTMDL~\cite{yang2014unified} and Ba {\em et al.}~\cite{lei2015predicting} fuse visual space and semantic space to a common intermediate space and then use a hinge ranking loss or a binary cross entropy loss (see Fig.~\ref{analysis1}(b)). For both groups, the learned embedding model will make the variance of $\mathbf{WA}$ to be smaller than that of $\mathbf{B}$, which would thus make the hubness problem worse. In summary, the hubness will persist regardless what embedding model is adopted, as long as  NN search is conducted in a high dimensional space. Our model does not worsen it, whist other deep models do, which leads to the performance difference as demonstrated in our experiments.

\section{Experiments}
\label{sec:exp}
\subsection{Dataset and settings}
\noindent
We follow two ZSL settings: the \emph{old} setting and the new \emph{GBU} setting provided by~\cite{xian2017zero} for training/test splits. 
Under the \emph{old} setting, adopted by most existing ZSL works before \cite{xian2017zero},
some of the test classes also appear in
the ImageNet 1K classes, which have been used to pretrain the image embedding 
network, thus violating the zero-shot assumption.
In contrast, the new \emph{GBU} setting ensures that none of the test classes of the datasets appear in
the ImageNet 1K classes. Under both settings, the test set can comprise only the unseen class samples (conventional test set setting) or a mixture of seen and unseen class samples. The latter, termed   generalised zero-shot learning (GZSL), is more realistic in practice.  

\noindent \textbf{Datasets} \quad Four benchmarks are selected for the \emph{old} setting: \textbf{AwA} (Animals with Attributes)~\cite{lampert2014attribute} consists of 30,745 images of 50 classes. It has a fixed split for evaluation with 40 training classes and 10 test classes. \textbf{CUB} (CUB-200-2011)~\cite{wah2011multiclass} contains 11,788 images of 200 bird species. We use the same split as in~\cite{akata2015evaluation} with 150 classes for training and 50 disjoint classes for testing. \textbf{ImageNet (ILSVRC) 2010 1K} \cite{ilsvrc} consists of 1,000 categories
and more than 1.2 million images. We use the same training/test split as ~\cite{mensink2012metric, frome2013devise} which gives 800 classes for training and 200 classes for testing. \textbf{ImageNet (ILSVRC) 2012/2010}: for this dataset, we use the same setting as~\cite{fu2016semi}, that is,  ILSVRC 2012 1K is used as the training seen classes, while 360 classes in ILSVRC 2010 which do not appear in ILSVRC 2012 are used as the test unseen classes.
Three datasets~\cite{xian2017zero} are selected for \emph{GBU} setting: \textbf{AwA1}, \textbf{AwA2} and \textbf{CUB}. The newly released AwA2~\cite{xian2017zero} consists of 37,322 images of 50 classes which is an extension of AwA while AwA1 is same as AwA but under the \emph{GBU} setting.

\noindent \textbf{Semantic space}\quad For \textbf{AwA}, we use the continuous 85-dimension class-level attributes provided in \cite{lampert2014attribute}, which have been used by all recent works.  For the word vector space, we use the 1,000 dimension word vectors provided in \cite{fu2014transductive, fu2015transductive}. For \textbf{CUB}, continuous 312-dimension class-level attributes and 10 descriptions per image provided in~\cite{reed2016learning} are used. For \textbf{ILSVRC 2010} and \textbf{ILSVRC 2012}, we trained a skip-gram language model~\cite{mikolov2013efficient, mikolov2013distributed} on a corpus of 4.6M Wikipedia documents to extract $1,000$D word vectors for each class. 

\noindent \textbf{Model setting and training} \quad Unless otherwise specified, We use the Inception-V2~\cite{szegedy2015going, ioffe2015batch} as the CNN subnet in the old and conventional setting, and ResNet101~\cite{he2016deep} for the GBU and generalised setting, taking the top pooling units as image embedding with dimension $D=1024$ and $2048$ respectively. The  CNN subnet is pre-trained on ILSVRC 2012 1K classification without fine-tuning, the same as the recent deep ZSL works~\cite{lei2015predicting, reed2016learning}. 
For fair comparison with DeViSE~\cite{frome2013devise}, ConSE~\cite{norouzi2013zero} and AMP~\cite{fu2015zero} on ILSVRC 2010, we also use the Alexnet~\cite{krizhevsky2012imagenet} architecture and  pretrain it from scratch  using the 800 training classes. All input images are resized to $224 \times 224$. Fully connected layers of our model are initialised with random weights for all of our experiments. Adam~\cite{kingma2014adam} is used to optimise our model with a learning rate of 0.0001 and a minibatch size of 64. The model is implemented based on {\em Tensorflow}.

\noindent \textbf{Parameter setting} \quad In the semantic encoding branch of our network, the output size of the first FC layer $M$ is set to 300 and 700 for AwA and CUB respectively when a single semantic space is used (see Fig.~\ref{fig:full_model}(a)). Specifically, we use one FC layer for ImageNet in our experiments. For multiple semantic space fusion, the multi-modal fusion layer output size is set to  900 (see Fig.~\ref{fig:full_model}(b)). 
When the semantic representation was encoded from descriptions for the CUB dataset, a bidirectional LSTM encoding subnet is employed (see Fig.~\ref{fig:full_model}(c)). We use the \verb|BasicLSTMCell| in {\em Tensorflow} as our RNN cell and employ ReLU as activation function. We set the input sequence length to 30; longer text inputs are cut off at this point and shorter ones are zero-padded. The word embedding size and the number of LSTM unit are both 512. Note that with this LSTM subnet, RMSprop is used in the place of Adam to optimise the  whole network with a learning rate of 0.0001, a  minibatch size of 64 and gradient clipped at 5. The loss weighting factor $\lambda$ in Eq.~(\ref{loss_function}) is searched by five-fold cross-validation. Specifically, 20\% of the seen classes in the training set are used to form a validation set.

\subsection{Experiments on small scale datasets}

\noindent \textbf{Competitors} \quad Numerous existing works reported results on AwA and CUB these two relatively small-scale datasets under old setting. Among them, only the most competitive ones are selected for comparison due to space constraint. The selected 13 can be categorised into the non-deep model group and the deep model group. All the non-deep models use ImageNet pretrained CNN to extract visual features. They differ in which CNN model is used: $F_O$ indicates that overfeat \cite{sermanet2013overfeat} is used; $F_G$ for GoogLeNet \cite{szegedy2015going}; and $F_V$ for VGG net \cite{simonyan2014very}. The second group are all neural network based with a CNN subnet. For fair comparison, we implement the models in \cite{frome2013devise, socher2013zero, yang2014unified, lei2015predicting} on AwA and CUB with Inception-V2 as the CNN subnet as in our model and \cite{reed2016learning}. The compared methods also differ in the semantic spaces used. Attributes (A) are used by all methods; some also use word vector (W) either as an alternative to attributes, or in conjunction with attributes (A+W). For CUB, recently the instance-level sentence descriptions (D) are used \cite{reed2016learning}. Note that only inductive methods are considered. Some recent methods ~\cite{zhang2016zeroshot, fu2014transductive, fu2015transductive} are tranductive in that they use all test data at once for model training, which gives them a big unfair advantage. 

\noindent \textbf{Comparative results on AwA under old setting} \quad
 From Table \ref{tab:awa_cub} we can make the following observations: (1) Our model DEM achieves the best results either with attribute or word vector. When both semantic spaces are used, our result is further improved to 88.1\%, which is  7.6\% higher than the best result reported so far \cite{zhang2016zero}.  (2) The performance gap between our model to the existing neural network based models are particularly striking. In fact, the four models \cite{frome2013devise,socher2013zero,yang2014unified,lei2015predicting} achieve weaker results than most of the compared non-deep models that use deep features only and do not perform end-to-end training. This verify our claim that selecting the appropriate visual-semantic embedding space is critical for the deep embedding models to work.  (3) As expected, the word vector space is less informative than the attribute space (86.7\% vs. 78.8\%) even though our word vector space alone result already beats all published results except for one \cite{zhang2016zero}.  Nevertheless, fusing the two spaces still brings some improvement (1.4\%).

\noindent \textbf{Comparative results on CUB under old setting} \quad
Table \ref{tab:awa_cub} shows that on the fine-grained dataset CUB, our model also achieves the best result. In particular, with attribute only, our result of 58.3\% is 3.8\% higher than the strongest competitor \cite{changpinyo2016synthesized}. The best result reported so far, however, was obtained by the neural network based 
 DS-SJE~\cite{reed2016learning} at 56.8\% using sentence descriptions. It is worth pointing out that this result was obtained using a word-CNN-RNN neural language model, whilst our model uses a bidirectional LSTM subnet, which is easier to train end-to-end with the rest of the network. When the same LSTM based neural language model is used, DS-SJE reports a lower accuracy of 53.0\%. Further more, with attribute only, the result of DS-SJE (50.4\%) is much lower than ours. This is significant because annotating attributes for fine-grained classes is probably just about manageable; but annotating 10 descriptions for each images is unlikely to scale to large number of classes. It is also evident that fusing attribute with descriptions leads to further improvement.

\setlength{\tabcolsep}{10pt}
\begin{table}[htbp]
\begin{center}
\footnotesize
\begin{tabular}{@{} l|c|c|cc @{}}
\toprule

\textbf{Model} & \textbf{F}     & \textbf{SS}   & \textbf{AwA} & \textbf{CUB} \\ 
\midrule 

AMP~\cite{fu2015zero}&$F_{O}$ & A+W & 66.0 & -  \\ 
SJE~\cite{akata2015evaluation}&$F_{G}$ & A & 66.7 &50.1  \\
SJE~\cite{akata2015evaluation}&$F_{G}$ &  A+W & 73.9 & 51.7 \\ 
ESZSL~\cite{romera2015embarrassingly}&$F_{G}$ & A & 76.3 & 47.2 \\
SSE-ReLU~\cite{zhang2015zero}&$F_{V}$ & A & 76.3 & 30.4 \\
JLSE~\cite{zhang2016zero}&$F_{V}$  & A & 80.5 & 42.1 \\ 
SS-Voc \cite{fu2016semi}&$F_{O}$ &A/W & 78.3/68.9 & - \\ 
SynC-struct~\cite{changpinyo2016synthesized}&$F_{G}$ & A & 72.9 & 54.5 \\ 
SEC-ML~\cite{bucher2016improving}&$F_{V}$  & A & 77.3 & 43.3 \\ 
\midrule
%\parbox{2mm}{\multirow{5}{*}[-3pt]{\rotatebox{90}{Deep}}}
DeViSE~\cite{frome2013devise}&$N_{G}$ & A/W &56.7/50.4  &33.5   \\ 
Socher {\em et al.}~\cite{socher2013zero}&$N_{G}$ & A/W & 60.8/50.3 &39.6  \\ 
MTMDL~\cite{yang2014unified}&$N_{G}$ & A/W &63.7/55.3   &32.3   \\ 
Ba {\em et al.}~\cite{lei2015predicting}&$N_{G}$ & A/W &69.3/58.7  &34.0   \\ 
DS-SJE~\cite{reed2016learning}&$N_{G}$ & A/D & - & 50.4/\textbf{56.8}  \\ 
\midrule
\midrule
DEM&$N_{G}$ & A/W(D) & \textbf{86.7}/\textbf{78.8} &\textbf{58.3}/53.5 \\ 
\midrule
DEM&$N_{G}$ & A+W(D) &\bf 88.1 &\bf 59.0 \\ 
\bottomrule
\end{tabular}
\end{center}
\caption{Zero-shot classification accuracy (\%) comparison on AwA and CUB (hit@1 accuracy over all samples) under the old and conventional setting. SS: semantic space; A: attribute space; W: semantic word vector space; D: sentence description (only available for CUB). F: how the visual feature space is computed; For non-deep models: $F_O$ if overfeat \cite{sermanet2013overfeat} is used; $F_G$ for GoogLeNet \cite{szegedy2015going}; and $F_V$ for VGG net \cite{simonyan2014very}. For neural network based methods, all use Inception-V2 (GoogLeNet with batch normalisation)~\cite{szegedy2015going, ioffe2015batch} as the CNN subnet, indicated as $N_G$.}
\label{tab:awa_cub}
\end{table}

\noindent \textbf{Comparative results under the GBU setting} \quad
We follow the evaluation setting of \cite{xian2017zero}. We compare our model with 13 alternative ZSL models in Table~\ref{tab:gzsl}. We can see that on AwA1, AwA2 and aPY, the proposed model DEM is particularly strong under the more realistic GZSL setting measured using the harmonic mean (H) metric. In particular, DEM achieves state-of-the-art performance on AwA1, AwA2 and SUN under conventional setting with $68.4\%$, $67.1\%$ and $61.9\%$, outperforming alternatives by big margins.

\setlength{\tabcolsep}{6pt}
\begin{sidewaystable}[htbp]
\centering
\footnotesize
\resizebox{1\columnwidth}{!}{%
\begin{tabular}{@{} l|c|ccc|c|ccc|c|ccc|c|ccc|c|ccc @{}}
\toprule
&\multicolumn{4}{c|}{\bf AwA1}&\multicolumn{4}{c|}{\bf AwA2}&\multicolumn{4}{c|}{\bf CUB}&\multicolumn{4}{c|}{\bf aPY}&\multicolumn{4}{c}{\bf SUN}\\
\midrule
&\bf ZSL & \multicolumn{3}{c|}{\bf GZSL}&\bf ZSL & \multicolumn{3}{c|}{\bf GZSL} &\bf ZSL & \multicolumn{3}{c|}{\bf GZSL}&\bf ZSL & \multicolumn{3}{c|}{\bf GZSL}&\bf ZSL & \multicolumn{3}{c}{\bf GZSL}\\
Model &\textbf{T1} & \textbf{u} & \textbf{s} & \textbf{H}&\textbf{T1} & \textbf{u} & \textbf{s} & \textbf{H} & \textbf{T1} & \textbf{u} & \textbf{s} & \textbf{H} &\textbf{T1} & \textbf{u} & \textbf{s} & \textbf{H}&\textbf{T1} & \textbf{u} & \textbf{s} & \textbf{H}\\
\midrule 
{{DAP}}~\cite{lampert2014attribute}&44.1 & 0.0& 88.7 &0.0 &46.1 &0.0&84.7  &0.0&40.0 &1.7& 67.9 &3.3 &33.8 &4.8 & 78.3&9.0 &39.9 &4.2 &25.1 &7.2 \\ 
{{IAP}}~\cite{lampert2014attribute}&35.9 &2.1 &78.2 &4.1&35.9 &0.9& 87.6 &1.8 &24.0    & 0.2 &\bf 72.8& 0.4&36.6 &5.7& 65.6& 10.4 &19.4 &1.0& 37.8 &1.8   \\
{{ConSE}}~\cite{norouzi2013zero} &45.6 &0.4 &88.6 &0.8&44.5 &0.5& 90.6  &1.0&34.3 &1.6& 72.2 &3.1 &26.9 &0.0 &\textbf{91.2}& 0.0   &38.8 &6.8& 39.9& 11.6   \\ 
{{CMT}}~\cite{socher2013zero} &39.5 &8.4 &86.9 &15.3&37.9  &8.7  &89.0 &  15.9    &34.6  &4.7 &60.1 & 8.7 & 28.0 &10.9  &74.2& 19.0  &39.9 & 8.7 &28.0 &13.3  \\ 
{{SSE}}~\cite{zhang2015zero} &60.1 & 7.0&80.5 &12.9&61.0 &8.1& 82.5 &14.8&43.9 &8.5&46.9  &14.4&34.0 &0.2 &78.9& 0.4 &51.5 &2.1 &36.4 &4.0  \\ 
{{DeViSE}}~\cite{frome2013devise} & 54.2&13.4 &68.7 &22.4&59.7 &  17.1& 74.7 &27.8&52.0 & 23.8& 53.0 &32.8 &\bf 39.8 & 4.9& 76.9& 9.2 &56.5 &16.9 &27.4 &20.9 \\ 
{{SJE}}~\cite{akata2015evaluation} & 65.6&11.3 &74.6 &19.6 &61.9 &  8.0 &73.9 &14.4&53.9 & 23.5 &59.2&33.6&32.9 & 3.7& 55.7& 6.9&   53.7&14.7 &30.5 &19.8\\ 
{{LATEM}}~\cite{xian2016latent}&55.1 &7.3 &71.7 & 13.3&55.8 & 11.5 &77.3 &20.0 &49.3 & 15.2 &57.3 &24.0&35.2 &0.1 &73.0 &0.2 &55.3 &14.7& 28.8 &19.5   \\ 
{{ESZSL}}~\cite{romera2015embarrassingly} &58.2 &6.6 &75.6 &12.1 &58.6 &  5.9& 77.8& 11.0&53.9 &  12.6 &63.8&21.0&38.3 & 2.4 &70.1 &4.6  &54.5 &11.0& 27.9& 15.8  	\\ 
{{ALE}}~\cite{akata2016label}&59.9 &16.8 &76.1 & 27.5 &62.5 & 14.0& 81.8& 23.9  &54.9 & 23.7& 62.8& 34.4 &39.7 &4.6 &73.7 & 8.7  &58.1 &\bf 21.8  &33.1 &\bf 26.3  \\ 
{{SYNC}}~\cite{changpinyo2016synthesized}&54.0 &8.9 &87.3 & 16.2 &46.6 & 10.0& 90.5& 18.0 & 55.6 & 11.5& 70.9& 19.8& 23.9&7.4 &66.3 &13.3   &56.3 &7.9 &\textbf{43.3} &13.4   \\ 
{{SAE}}~\cite{kodirov2017semantic}&53.0 & 1.8&77.1 &3.5&54.1 &1.1  & 82.2 & 2.2 & 33.3 & 7.8 & 54.0 & 13.6 &8.3 &0.4 &80.9 &0.9 &40.3 &8.8 & 18.0&11.8 \\ 
{{Relation Net}}~\cite{sung2018learning}&68.2 & 31.4&\bf 91.3 &46.7 &64.2 &30.0  &\bf 93.4 &\bf 45.3 &\bf 55.6 &\bf 38.1 & 61.1 &\bf 47.0 &- &- &- &- &- &- & -&- \\ 
\midrule
{{DEM}}&\bf 68.4 &\bf 32.8 &84.7 &\bf 47.3&\bf 67.1  &\bf 30.5  &86.4  & 45.1 &51.7 &19.6 &57.9 &29.2 &35.0 &\bf 11.1 &75.1 &\bf 19.4 &\bf 61.9 &20.5 &34.3 &25.6 \\ 
\bottomrule
\end{tabular}%
}
\vspace{0.2cm}
\caption{\footnotesize Comparative results on four datasets. Under that 
ZSL setting, the performance is evaluated using  per-class average Top-1 (\textbf{T1}) accuracy (\%), and under GZSL, it is  measured using  \textbf{u} = \textbf{T1} on unseen classes, \textbf{s} = \textbf{T1} on seen classes, and \textbf{H} = harmonic mean.}
\label{tab:gzsl}
\end{sidewaystable}

\subsection{Experiments on ImageNet}
\noindent \textbf{Comparative results on ILSVRC 2010} \quad Compared to AwA and CUB, far fewer works report results on the large-scale ImageNet ZSL tasks. We compare our model against 8 alternatives on ILSVRC 2010 in Table \ref{tab:ILSVRC2010}, where we use hit@5 rather than hit@1 accuracy as in the small dataset experiments. Note that existing works follow two settings.  
Some of them \cite{Mukherjee2016Gaussian, Huang2013Local} use existing CNN model (e.g.~VGG/GoogLeNet) pretrained from ILSVRC 2012 1K classes to initialise their model or extract deep visual feature. Comparing to these two methods under the same setting, our model gives 60.7\%, which  beats the nearest rival PDDM~\cite{Huang2013Local} by over 12\%.
For comparing with the other 6 methods, we follow their setting and  pretrain our CNN subnet from scratch with Alexnet~\cite{krizhevsky2012imagenet} architecture using the 800 training classes for fair comparison. The results show that again, significant improvement has been obtained with our model.

\setlength{\tabcolsep}{10pt}
\begin{table}[htbp]
\begin{center}
\footnotesize
\begin{tabular}{@{} l |c@{}}
\toprule
{\textbf{Model}} &\textbf{hit@5}  \\ 
\midrule 
ConSE~\cite{norouzi2013zero} &28.5  \\ 
DeViSE~\cite{frome2013devise} & 31.8  \\ 
Mensink {\em et al.}~\cite{mensink2012metric} &35.7   \\
Rohrbach~\cite{rohrbach2011evaluating} &34.8 \\ 
PST~\cite{rohrbach2013transfer} &34.0 \\ 
AMP~\cite{fu2015zero} &41.0 \\ 
\midrule
Ours     &\bf 46.7 \\ 
\bottomrule
\toprule
Gaussian Embedding~\cite{Mukherjee2016Gaussian}&45.7 \\ 
PDDM~\cite{Huang2013Local}&48.2 \\ 
\midrule
DEM     &\bf 60.7  \\ 
\bottomrule
\end{tabular}
\end{center}
\caption{Comparative results (\%) on ILSVRC 2010 (hit@1 accuracy over all samples) under the old and conventional setting. }
\label{tab:ILSVRC2010}
\end{table}

\noindent \textbf{Comparative results on ILSVRC 2012/2010} \quad 
Even fewer published results on this dataset are available. Table \ref{tab:ILSVRC2012}   shows that our model clearly outperform the state-of-the-art alternatives by a large margin.

\setlength{\tabcolsep}{10pt}
\begin{table}[htbp]
\begin{center}
\footnotesize
\begin{tabular}{@{} l|c|c c@{}}
\toprule
{\textbf{Model}} &\textbf{hit@1} &\textbf{hit@5} \\ 

\midrule 
ConSE~\cite{norouzi2013zero} & 7.8 & 15.5 \\ 
DeViSE~\cite{frome2013devise} & 5.2 & 12.8 \\ 
AMP~\cite{fu2015zero} &6.1 &13.1\\ 
SS-Voc~\cite{fu2016semi} &9.5 & 16.8 \\ 

\midrule
DEM  &\bf 11.0 &\bf 25.7 \\ 
\bottomrule
\end{tabular}
\end{center}
\caption{Comparative results (\%) on  ILSVRC 2012/2010 (hit@1 accuracy over all samples) under the old and conventional setting. }
\label{tab:ILSVRC2012}
\end{table}

\subsection{Further analysis}

\begin{figure*}[th]
\centering
\begin{tabular}{cc}
{\includegraphics[width=6.55cm]{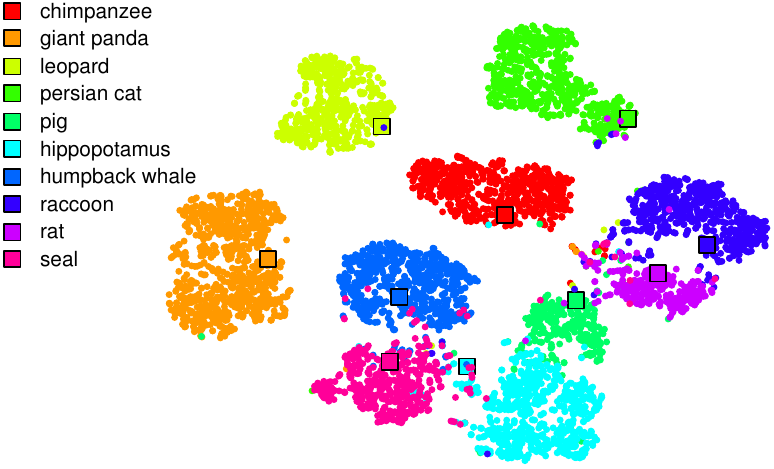}}&
{\includegraphics[width=5cm]{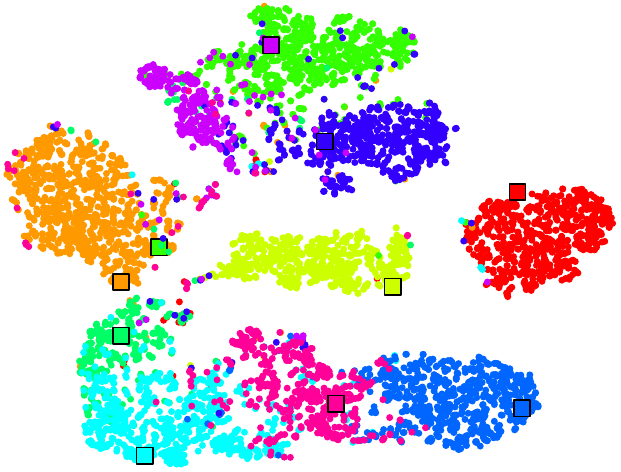}}\\
(a) S $\rightarrow$ V &(b) V $\rightarrow$ S
\end{tabular}
\vspace{0.3cm}
\caption{Visualisation of the distribution of the 10 unseen class images in the two embedding spaces on AwA using t-SNE~\cite{maaten2008visualizing}. Different classes as well as their corresponding class prototypes (in squares) are shown in different colours. Better viewed in colour. }
\label{fig:hubness_real}
\end{figure*}

\noindent \textbf{Importance of embedding space selection} \quad We argued that the key for an effective deep embedding model is the use of the CNN output visual feature space rather than the semantic space as the embedding space. In this experiment, we modify our model in Fig.~\ref{fig:full_model} by moving the two FC layers from the semantic embedding branch to the CNN feature extraction branch so that the embedding space now becomes the semantic space (attributes are used). Table \ref{tab:inverse_mapping} shows that by mapping the visual features to the semantic embedding space, the performance on AwA drops by 26.1\% on AwA, highlighting the importance of selecting the right embedding space. We also hypothesised that using the CNN visual feature space as the embedding layer would lead to less hubness problem. To verify that we measure the hubness using the skewness score (see Sec.~\ref{sec:hubness}). Table \ref{tab:hubness_score} shows clearly that the hubness problem is much more severe when the wrong embedding space is selected. We also plot the data distribution of the 10 unseen classes of AwA together with the prototypes. Figure \ref{fig:hubness_real} suggests that with the visual feature space as the embedding space, the 10 classes form compact clusters and are near to their corresponding prototypes, whilst in the semantic space, the data distributions of different classes are much less separated  and a few prototypes are clearly hubs causing miss-classification.

\setlength{\tabcolsep}{10pt}
\begin{table}[htbp]
\begin{center}
\footnotesize
\begin{tabular}{@{} l|c|c @{}}
\toprule
\textbf{Loss} &\textbf{Visual $\rightarrow$ Semantic} &\textbf{Semantic $\rightarrow$ Visual}  \\ 
\midrule 
Least square loss &60.6 &\bf 86.7  \\ 
Hinge loss&57.7 &72.8   \\ 
\bottomrule
\end{tabular}
\end{center}
\caption{Effects of selecting different embedding space and different  loss functions on zero-shot classification accuracy (\%) on AwA.}
\label{tab:inverse_mapping}
\end{table}

\setlength{\tabcolsep}{10pt}
\begin{table}[htbp]
\begin{center}
\footnotesize
\begin{tabular}{@{} l|c|c @{}}
\toprule
\textbf{ $N_{1}$ skewness} &\textbf{AwA} &\textbf{CUB}  \\ 
\midrule 
Visual $\rightarrow$ Semantic &0.4162 & 8.2697  \\ 
Semantic $\rightarrow$ Visual &\bf -0.4834 &\bf  2.2594  \\ 

\bottomrule
\end{tabular}
\end{center}
\caption{$N_{1}$ skewness score on AwA and CUB with different embedding space.}
\label{tab:hubness_score}
\end{table}

\noindent \textbf{Neural network formulation} \quad
Can we apply the idea of using visual feature space as embedding space to other models? To answer this, we consider a very simple model based on linear ridge regression which maps from the CNN feature space to the attribute semantic space or vice versa. In Table \ref{tab:linear_regression}, we can see that even for such a simple model, very impressive results are obtained with the right choice of embedding space. The results also show that with our neural network based model, much better performance can be obtained due to the introduced nonlinearity and its ability to learn end-to-end.  

\setlength{\tabcolsep}{10pt}
\begin{table}[htbp]
\begin{center}
\footnotesize
\begin{tabular}{@{} l|c|c @{}}
\toprule
\textbf{Model} &\textbf{AwA} &\textbf{CUB}  \\ 
\midrule 
Linear regression (V $\rightarrow$ S)&54.0 & 40.7 \\ 
Linear regression (S $\rightarrow$ V)&74.8 & 45.7  \\ 
DEM &\bf 86.7 &\bf 58.3   \\ 
\bottomrule
\end{tabular}
\end{center}
\caption{Zero-shot classification accuracy (\%) comparison with linear regression on AwA and CUB.}
\label{tab:linear_regression}
\end{table}

\noindent \textbf{Choices of the loss function} \quad As reviewed in Sec.~\ref{sec:related work}, most existing ZSL models use either margin based losses or binary cross entropy loss to learn the embedding model. In this work, least square loss is used. Table \ref{tab:inverse_mapping} shows that when the semantic space is used as the embedding space, a slightly inferior result is obtained using a hinge ranking loss in place of least square loss in our model. However, least square loss is clearly better when the visual feature space is the embedding space.

\section{Conclusion}
We have proposed a novel deep embedding model for zero-shot learning. The model differs from existing ZSL model in that it uses the CNN output feature space as the embedding space. We hypothesise that this embedding space would lead to less hubness problem compared to the alternative selections of embedding space. Further more, the proposed model offers the flexible of utilising multiple semantic spaces and is capable of end-to-end learning when the semantic space  itself is computed using a neural network. Extensive experiments show that our model achieves state-of-the-art performance on a number of benchmark datasets and validate the hypothesis that selecting the correct embedding space is the key for achieving the excellent performance. 

\section*{Acknowledgement}
This work was funded in part by the European FP7 Project SUNNY (grant agreement no. 313243).

{\small
\bibliographystyle{ieee}
\bibliography{egbib}
}

\end{document}